\documentclass[a4paper, 10pt, conference]{ieeeconf}      

\IEEEoverridecommandlockouts                              

\usepackage{cite}
\usepackage{amsmath,amssymb,amsfonts}
\usepackage{mathtools}
\usepackage{algorithm}
\usepackage{algpseudocode}

\usepackage{graphicx}
\usepackage{textcomp}
\usepackage{xcolor}

\usepackage{tikz}
\usepackage{pgfplots} 
\pgfplotsset{compat=newest}
\usetikzlibrary{spy,arrows,positioning,shapes,shapes.geometric,external,shadows,babel,calc, backgrounds}

\usepackage{booktabs}
\usepackage{diagbox}
\usepackage{siunitx}
\usepackage{subfigure}
\usepackage{makecell}

\makeatletter
\let\NAT@parse\undefined
\makeatother
\usepackage{hyperref}  

\pgfplotsset{
	colormap={radar_anomalies}{color(0)=(blue), color(1)=(orange), color(2)=(red), color(3)=(blue)}
}

\pgfplotsset{
	colormap={radar_anomalies_light}{color(0)=(lightgray) color(1)=(orange), color(2)=(red), color(4)=(blue)}
}

\pgfplotsset{
	colormap={radar_qualitative}{color(0)=(blue) color(1)=(green), color(2)=(red), color(3)=(orange), color(4)=(blue)}
}

\newcommand{\DrawBox}[6]{
\draw[#1, rotate around={#2: (axis cs: #3, #4)}]  (axis cs: #3, #4) rectangle (axis cs: #3 - #5, #4 + #6);
}

\newcommand{\VisualizeRadar}[2] {
	\addplot[scatter, only marks,
	scatter src=explicit symbolic,
	scatter/classes={#2},
	visualization depends on={\thisrow{rcs} \as \perpointmarksize},			
	scatter/@pre marker code/.append style={
		/tikz/mark size={(25pt+\perpointmarksize/1.5)/20}
	}
	]
	table[x=y,y=x, meta=label, col sep=comma] {#1};
	
	\addplot[point meta=\thisrow{label}, 
	quiver={
		u=\thisrow{vd_u},
		v=\thisrow{vd_v},
		scale arrows=1,
		every arrow/.append style={
			color=mapped color
		},
	},
	-stealth,
	] 
	table[x=y, 					
	y=x, 						
	meta=label, 
	col sep=comma, 		
	restrict expr to 
	domain={\thisrow{v_comp_abs}}{1:+inf}]	
	{#1};
}

\newcommand{\uproman}[1]{\uppercase\expandafter{\romannumeral#1}}

\newcommand\copyrighttext{\footnotesize \textcopyright~2021 IEEE. Personal use of this material is permitted. Permission from IEEE must be obtained for all other uses, in any current or future media, including reprinting/republishing this material for advertising or promotional purposes, creating new collective works, for resale or redistribution to servers or lists, or reuse of any copyrighted component of this work in other works.
}%

\newcommand\copyrightnotice{%
	\begin{tikzpicture}[remember picture,overlay]
	\node[anchor=south,xshift=9pt,yshift=10pt] at (current page.south) {\fbox{\parbox{\dimexpr\textwidth-\fboxsep-\fboxrule\relax}{\copyrighttext}}};
	\end{tikzpicture}%
}

\title{\LARGE \bf
Anomaly Detection in Radar Data Using PointNets
}

\author{Thomas Griebel\textsuperscript{*}, Dominik Authaler\textsuperscript{*}, Markus Horn\textsuperscript{*}, Matti Henning, Michael Buchholz, and Klaus Dietmayer
\thanks{All authors are with the Institute of Measurement, Control and Microtech-nology,  
	Ulm  University,  Albert-Einstein-Allee  41,  89081  Ulm,  Germany {\tt\small \{firstname\}.\{lastname\}@uni-ulm.de}}%
\thanks{\textsuperscript{*}Thomas Griebel, Dominik Authaler and  Markus Horn are co-first authors. Corresponding author: Thomas Griebel.}%
\thanks{This work was supported by the projects SecForCARs (reference number: 16KIS0795) and UNICARagil (reference number: 16EMO0290).
We acknowledge the financial support for the projects by the Federal Ministry of Education and Research of Germany (BMBF).
}
}

\begin{document}

\maketitle%
\thispagestyle{empty}
\pagestyle{empty}

\begin{abstract}
For autonomous driving, radar is an important sensor type. 
On the one hand, radar offers a direct measurement of the radial velocity of targets in the environment. 
On the other hand, in literature, radar sensors are known for their robustness against several kinds of adverse weather conditions. 
However, on the downside, radar is susceptible to ghost targets or clutter which can be caused by several different causes, e.g., reflective surfaces in the environment. 
Ghost targets, for instance, can result in erroneous object detections. 
To this end, it is desirable to identify anomalous targets as early as possible in radar data.
In this work, we present an approach based on PointNets to detect anomalous radar targets. 
Modifying the PointNet-architecture driven by our task, we developed a novel grouping variant which contributes to a multi-form grouping module. 
Our method is evaluated on a real-world dataset in urban scenarios and shows promising results for the detection of anomalous radar targets.
\end{abstract}

\section{Introduction}

\copyrightnotice%

Nowadays, radar sensors are widely used for modern advanced driver assistance systems.
In contrast to other used sensor types such as camera and lidar, radar is known for its robustness regarding adverse weather conditions because of its comparatively large wavelength of about \SI{4}{mm} for a \SI{77}{GHz} radar.
Moreover, radar is able to directly measure the radial velocity of an object through the Doppler effect. 
Other sensors usually need at least measurements from two time steps to provide a velocity estimation.
On the downside, radar data are sparse and prone to aspects like ghost targets. 
Some more challenges that radar has to deal with are noise, interference between radar sensors, measurement ambiguities, and multi-path propagation.

These challenging aspects in radar have gained more and more attention in literature in recent years. 
In~\cite{kraus2020using}, Kraus et al.~present a method to segment radar targets into real and ghost objects.
However, only anomalies caused by multi-path propagation are considered with a focus on related certain situations in their dataset. 
In addition to that, they examined only vulnerable road users, such as bicyclists and pedestrians, for ghost objects detection. 
For the evaluation, a dual-sensor setup is used where measurements are combined and accumulated over \SI{200}{\milli\second}.
A similar task is approached in~\cite{roos2017ghost} where the authors propose a model-based detection algorithm for anomalies. 
This method needs two radar sensors and is limited to the detection of multi-path-related anomalies which cause ghost objects.
In~\cite{prophet2019instantaneous}, a detection algorithm consisting of three consecutive steps using handcrafted features is presented in order to detect anomalies. 
The used dataset is limited to targets whose Doppler velocity is within the unambiguously measurable range.
However, all of these works have at least one of the following restrictions: focusing solely on multi-path anomalies, accumulating multiple sensors and measurements over time, detecting anomalies at object level and not at target level, and using handcrafted features where expert knowledge is needed.
Although handcrafted features are easier to interpret in general, our radar sensor is a black box where expert knowledge is missing.

\begin{figure}[!t]
    \centerline{
        \hspace{0.9cm}
		\subfigure{
		    \includegraphics[width=0.75\linewidth]{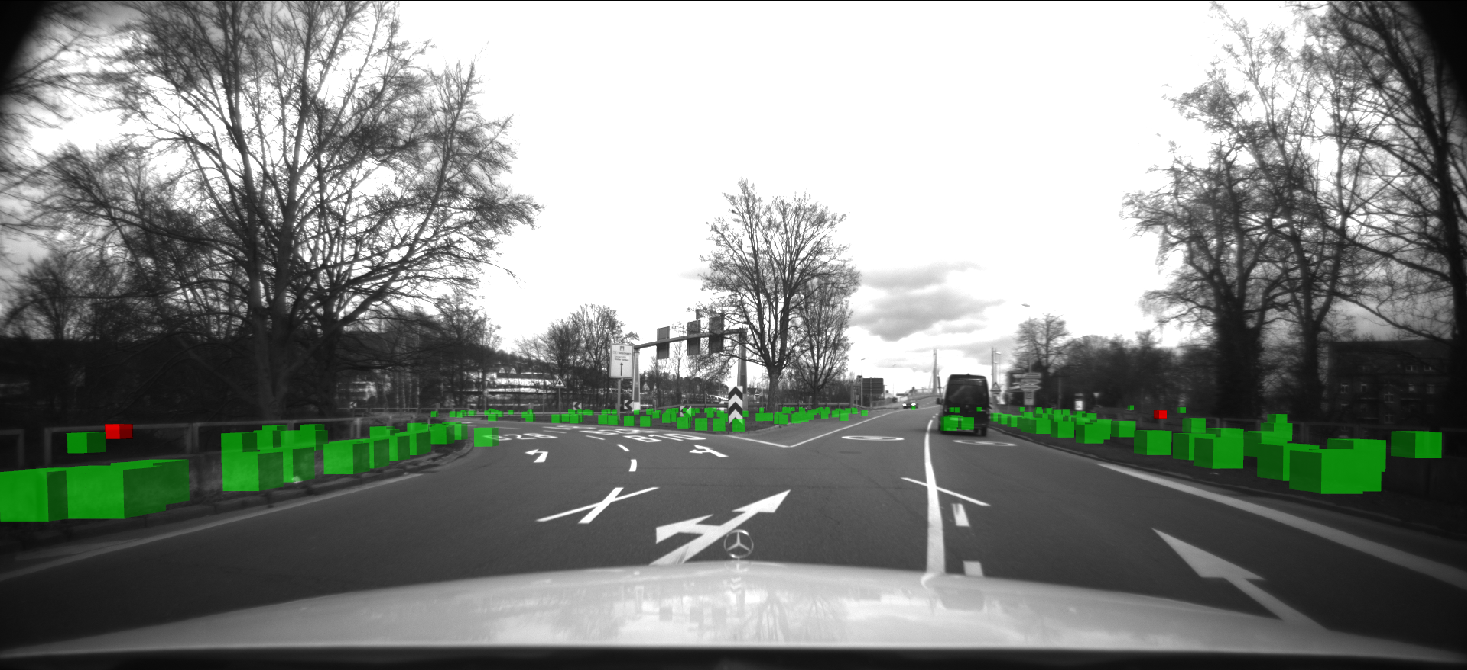}
		}
	}
	\centerline{
		\subfigure{
		    \begin{tikzpicture}
		        \begin{axis}[
        			width=0.95*\linewidth, 			
        			grid=major, 					
        			grid style={dashed,gray!30}, 	
        			xlabel= $y$ in \si{\meter}, 	
        			ylabel= $x$ in \si{\meter},
        			x dir=reverse,
        			axis equal image,
        			xmin=-45,
        			xmax=45,
        			ymax=48,
        			ymin=-7.5,
        			colormap={CM}{
        				samples of colormap=(4 of radar_anomalies)},
        			colormap access=piecewise constant,
        			point meta min=0, point meta max=4,
            		legend columns = 3,
            		legend style={fill=white, fill opacity=0.8, draw opacity=1.0, text opacity=1.0, at={(0.925,0.1875),anchor=south}},
            		/tikz/every even column/.append style={column sep=0.3cm},
            		legend cell align={left}
            		]        			\VisualizeRadar{resources/figures/angular_anomaly_fc_25_with_sensor_calib_applied.csv}{
        				0={mark=*, black},
        				4={mark=*, blue},
        				2={mark=*, red},
        				1={mark=*, orange},
        				3={mark=*, black}			
        			}
        			\addlegendentry{stationary}
        			\addlegendentry{moving}
        			\addlegendentry{anomalous}
        			\draw[line width=0.35mm, gray, rotate around={-16: (axis cs: -7.1, 25.5)}]  (axis cs: -7.1, 25.5) rectangle (axis cs:-9.5, 31);
		        \end{axis}
		    \end{tikzpicture}
		}
	}
	\vspace{-0.25cm}
	\caption{Exemplary illustration of anomalous radar targets. Radar data are visualized below the corresponding camera image. The red points display the anomalous radar targets, which show a significant Doppler velocity, although no moving objects are perceivable. The ego-motion compensated Doppler velocity is visualized by the length of the arrows. The gray bounding box sketches the preceding vehicle.\label{fig:eye_catcher}}
	\vspace{-0.5cm}
\end{figure}
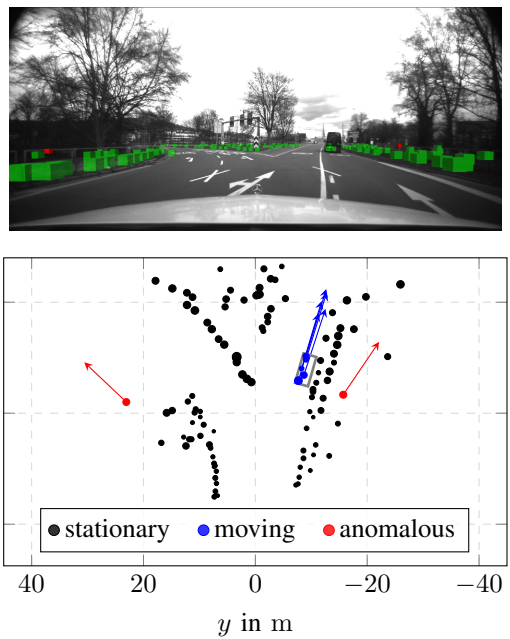
In this work, we present a single-shot approach to identify anomalous radar targets using deep learning methods. 
We consider the radar data of a single measurement cycle as a point cloud, which we call radar target list. 
Each radar target consists of two spatial coordinates in combination with the ego-motion compensated and uncompensated Doppler velocity and the radar cross section (RCS) as input features. 
In Fig.~\ref{fig:eye_catcher}, exemplary radar data including some anomalous radar targets are visualized. 
Since radar targets are represented as point cloud, we leverage PointNets~\cite{PointNet,PointNet++} which are able to directly process point clouds. 
We modify the PointNet-architecture by a novel grouping variant which is tailor-made for the anomaly detection task and contributes to a multi-form grouping module.

Our main contributions are:
\begin{itemize}
	\item a characterization of sensor-specific anomalies in radar in Section~\ref{section:sensor_setup},
	\item a single-shot anomaly detector in radar data in Section~\ref{section:anomaly_detection},
	\item a novel multi-form grouping module driven by radar anomaly characteristics in Section~\ref{section:mulit_form_group},
	\item an extensive evaluation on real-world data in Section~\ref{section:experiments}.
\end{itemize}

\section{Sensor Setup} \label{section:sensor_setup}

In this work, we use the ARS 408-21 Long Range Radar \SI{77}{GHz} Premium (ARS 408-21) sensor, which is an industrial sensor developed by Continental AG. 
The ARS~400 series of radar sensors was initially developed for the automotive industry and is widely used for automotive applications such as advanced driver assistance systems~\cite{weber2020automotive}.
As already mentioned, we consider radar data as a point cloud consisting of many radar targets. 
Exactly these data are supplied by the interface of the ARS 408-21. 
Thus, we consider the sensor as a black box because we are not able to get insights into the signal processing of the raw radar data. 
After the non-transparent signal processing, the sensor outputs resolved radar reflections which we call radar targets. 

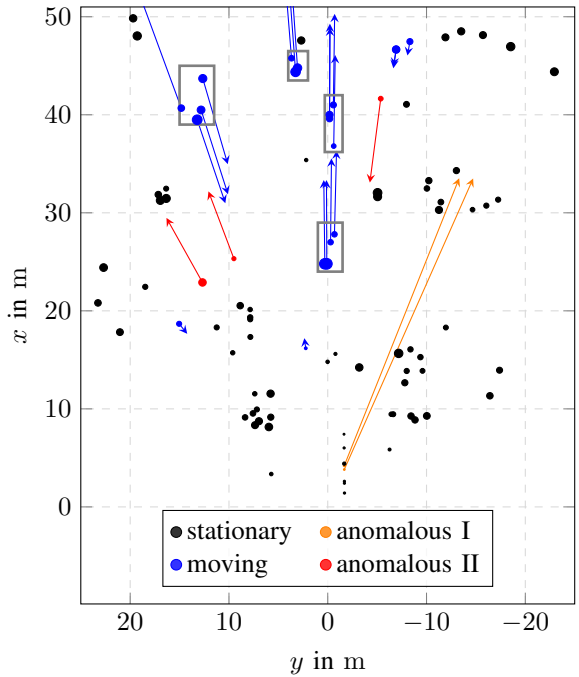
\begin{figure}[t]
	\centering
	\begin{tikzpicture}
		\begin{axis}[
		width=1.275\linewidth, 			
		grid=major, 					
		grid style={dashed,gray!30}, 	
		xlabel= $y$ in \si{\meter}, 	
		ylabel= $x$ in \si{\meter},
		x dir=reverse,
		axis equal image,
		xmin=-25,
		xmax=25,
		ymax=51,
		ymin=-9.9,
		colormap={CM}{
			samples of colormap=(4 of radar_anomalies)},
		colormap access=piecewise constant,
		point meta min=0, point meta max=4,
		legend style={fill=white, fill opacity=0.8, draw opacity=1.0, text opacity=1.0, at={(0.835,0.1575),anchor=south}},
		/tikz/column 2/.style={
                column sep=0.3cm,
            },
		legend cell align={left},
		legend columns = 2
		]
		\VisualizeRadar{resources/figures/more_anomalies_fc_1472_with_sensor_calib_applied.csv}{
			0={mark=*, black},
			1={mark=*, orange},
			4={mark=*, blue},
			2={mark=*, red},
			3={mark=*, black}
		}
		\addlegendentry{stationary}
		\addlegendentry{anomalous \uproman{1}}
		\addlegendentry{moving}
		\addlegendentry{anomalous \uproman{2}}
		\draw[line width=0.35mm, gray, rotate around={0: (axis cs: 1, 24)}]  (axis cs: 1, 24) rectangle (axis cs: -1.5, 29);
		\draw[line width=0.35mm, gray, rotate around={0: (axis cs: 15, 39)}]  (axis cs: 15, 39) rectangle (axis cs: 11.5, 45);
		\draw[line width=0.35mm, gray, rotate around={0: (axis cs: 0.3, 36.2)}]  (axis cs: 0.3, 36.2) rectangle (axis cs: -1.5, 42);
		\draw[line width=0.35mm, gray, rotate around={0: (axis cs: 2, 43.5)}]  (axis cs: 2, 43.5) rectangle (axis cs: 4, 46.5);
		\end{axis}
	\end{tikzpicture}
	\vspace{-0.25cm}
	\caption{Exemplary illustration of a radar measurement with several anomalous radar targets. Anomalies colored in red are probably related to errors in the direction of arrival estimation. Whereas anomalies colored in orange are probably related to the multi-path propagation effect or possibly to Doppler velocity ambiguities. The gray bounding boxes sketch vehicles in the environment.\label{fig:considered_anomalies}}
	\vspace{-0.5cm}
\end{figure}

Fig.~\ref{fig:considered_anomalies} provides a visualization of a radar measurement with five anomalous targets. 
These targets are either highlighted in orange or red to distinguish two different kinds of anomalies.
To give the reader an intuition about what can be seen in the radar data, cars are marked with gray boxes. 
However, these boxes are only meant for highlighting the radar targets on cars and do not claim to represent their exact bounding box. 
The length of the arrows visualizes the ego-motion compensated Doppler velocity of the target. 
For clarity purposes, arrows are only drawn for a compensated Doppler velocity greater than \SI{1}{m/s}. 
Finally, the size of the points represents the RCS value.

Due to the unknown sensor-specific signal processing procedure, some of the hereby considered anomalies are claimed to be ARS 408-21 specific. Hence, we claim that the sensor is not always able to filter out all anomalies in its signal processing. Moreover, some hypotheses, stating the reasons of the sensor-specific anomalies, are developed in the following.
The anomalies colored in red in Fig.~\ref{fig:considered_anomalies} are targets that are quite similar to car targets within the same range bin. 
Here, the ego-motion compensated Doppler velocity may vary compared to car targets on the same range. 
This is caused by the ego-motion compensation itself which takes the azimuth angle into account and, thus, motivates the additional consideration of the uncompensated Doppler velocity.
The high similarities in both the measured range and Doppler velocity of the anomalous targets lead to the hypothesis that the anomalies are caused by errors in the direction of arrival (DoA) estimation, also referred to as azimuth angle. 
These errors may be related to measurement ambiguities and may be caused by the sensor design, e.g., antenna designs or beamforming.

Furthermore, anomalies may also be related to multi-path propagation effects. Multi-path propagation describes a phenomenon, where the radar sensor receives echo signals which do not propagate along the direct line of sight between sensor and object. Instead, the propagation paths include reflections on other surfaces. 
The anomalies colored in orange in Fig.~\ref{fig:considered_anomalies} characterizes a high Doppler velocity surrounded by mostly stationary targets. These anomalies are probably related to multi-path propagation effects. In this case, a possible propagation path leading to the high measured Doppler velocity may include multiple bounces on the test vehicle. Due to the ego-motion of this vehicle, these bounces increase the frequency of the reflected wave and thus the measured Doppler velocity. 
However, this is just one possible explanation of the anomaly and, based on the measurement visualized in Fig.~\ref{fig:considered_anomalies}, we can neither prove nor debunk this hypothesis.
This means that other influences are conceivable that lead to anomalies in radar data such as Doppler velocity ambiguities.

Throughout this work, we focus on these kinds of anomalies, since they can be comprehensibly identified based on a single radar point cloud using a camera image for verification purposes.

\section{Related Work} \label{section:related_work}

As already mentioned in Section~\ref{section:sensor_setup}, multiple effects have to be considered for our presented anomalies.
Hence, approaches proposed in literature are diverse, aiming at different stages of the processing chain for radar data.
In~\cite{roos2018enhancement} a novel signal processing algorithm for an enhancement of the ambiguity range of the Doppler velocity measurement is proposed. 
In this way, anomalies resulting from measurement ambiguities are reduced. 
However, this does not affect the anomalies caused by multi-path propagation effects, which are, e.g.,
analyzed in~\cite{kaman2018automotive, roos2017ghost}. 
The authors of~\cite{roos2017ghost} additionally propose a model-based detection algorithm for the anomalies. 
This algorithm relies on a comparison of the current motion state and the estimated bounding box of a target vehicle. 
It is worth mentioning that the estimation of the motion state requires the usage of two radar sensors. 
Moreover, to estimate the bounding box correctly the vehicle has to be represented by multiple radar targets. 
Finally, the algorithm is limited to the detection of multi-path related anomalies that cause ghost objects and cannot detect single anomalous targets caused by factors such as measurement ambiguities. 

Other approaches for anomaly detection involve the usage of machine learning techniques, especially of deep learning. As presented in~\cite{chalapathy2019deep}, these techniques have been applied to many different anomaly detection tasks. 
However, most of the works focusing on the detection of radar anomalies consider this task as a semantic segmentation problem. 
Hence, we adopt this approach for our investigation.
The authors of~\cite{prophet2019instantaneous} present a detection algorithm suitable for the usage of a single measurement of one radar sensor. 
The algorithm is structured in three consecutive steps. First, moving targets are identified. 
In the second step, handcrafted features for the identified moving targets are calculated. These features are then passed to a random forest classifier in the final step. 
In contrast to the detection algorithm described in~\cite{roos2017ghost}, this algorithm is capable of detecting anomalies regardless of their causes. 
Furthermore, it shows promising results on the task of classifying targets in the radar point cloud into infrastructure, real moving targets, and anomalies. However, it is necessary to note that the used dataset is limited to targets whose Doppler velocity is within the unambiguously measurable range. Hence, errors in the first algorithm step are limited to the rare case of purely tangential moving targets. 
As a modification of this approach, \cite{garcia2019moving} presents a detection algorithm based on deep learning. 
Instead of handcrafted features, the algorithm uses an occupancy grid map and a map of moving targets as input. 
The latter incorporates the concept of the first stage of the algorithm presented in~\cite{prophet2019instantaneous}. Moreover, the random forest classifier is replaced by a convolutional neural network (CNN), which performs a segmentation of the input data. Considering only targets with a maximum longitudinal distance of \SI{30}{\meter}, the algorithm shows promising results. However, to use the radar point cloud in the algorithm described above, an occupancy grid map has to be calculated. As the authors of~\cite{schumann2018semantic} show, this step can be omitted. Instead, the task of semantic segmentation can be applied directly to radar point clouds using the PointNet++ architecture~\cite{PointNet++}. 
In order to overcome the sparsity of radar measurements, the point clouds provided by multiple radar sensors are combined and accumulated over \SI{500}{\milli\second}. Using this accumulated point cloud as input, the PointNet++ segmentation architecture is used to classify different classes of road users. 
In~\cite{chamseddine2021ghost}, the PointNet-architecture is used to detect ghost targets in 3D radar point clouds. In their work, ghost targets are referred to as multi-path reflections. 
It is noteworthy that highly dense radar data are used, which consist of around $1000$ 3D points per measurement, as opposed to around $200$ 2D points from our used radar sensor.
Based on the contribution of~\cite{schumann2018semantic}, Kraus et~al. present in~\cite{kraus2020using} an application of the PointNet++ architecture to the task of segmenting radar targets into real and ghost objects.
Although this task is similar to our considered task, there are also fundamental differences. 
The authors of~\cite{kraus2020using} use a dual-sensor setup, in which measurements are combined and accumulated over \SI{200}{\milli\second} to increase the density of the radar point cloud.
In addition to that, only vulnerable road users, such as bicyclists and pedestrians, and their corresponding ghost objects caused by multi-path propagation effects are considered. 
In contrast to that, our detector is able to derive information about anomalies from only a single radar point cloud. 
Moreover, we primarily consider vehicles and aim on detecting even single anomalous targets. 
The causes of the anomalies are also not restricted to multi-path propagation effects.

To summarize this section, to the best of our knowledge, we propose the first anomaly detector in 2D sparse radar data using deep learning methods which combines all of the following aspects, i.e., single-shot detection, consideration of various kinds of anomalies, and approaching the task on a target level.

\section{Problem Statement} \label{section:problem_statement}

The goal of our proposed method is to detect anomalies in radar data. 
In this work, anomalies are defined as radar targets with significant Doppler velocities which apparently do not correspond to real-world moving objects.
To identify these anomalous radar targets, radar data is given as point clouds. 
In detail, the radar point cloud $P$ consists of a set of five-dimensional points $P = \left\{p_i \in \mathbb{R}^5 | i = 1, \ldots, n \right\}$ with $n \in \mathbb{N}, n \leq 250$ the number of points per time step $k = 1, \ldots, m$.
Each point represents a radar target which is obtained by some sensor-processing of raw radar data. 
Each target can be described by $p_i = \left(x, y, \tilde{v}_D, \sigma, v_D \right)$ where $(x,y)$ denotes the 2D-position, $\tilde{v}_D$ and $v_D$ the ego-motion compensated and uncompensated Doppler velocity, and $\sigma$ the RCS of the target. 
For our anomaly detection task, radar data of only one single measurement cycle of one radar sensor are used. 
This means that our radar data are augmented neither by accumulating measurements over time nor by using data from multiple sensors.
Moreover, the output of our anomaly detection method is a binary segmentation of the radar point cloud where each target is classified either as normal (non-anomalous) or as anomalous.

\section{Anomaly Detection in Radar Data} \label{section:anomaly_detection}

In this section, we present our proposed method for the detection of anomalies in radar data.
After explaining the used architecture variants of PointNets, we focus on our developed multi-form grouping module.

\subsection{PointNets} \label{section:pointnets}

We investigate four different forms of PointNets~\cite{PointNet, PointNet++} for the previously described anomaly segmentation task. Namely, we evaluate one PointNet architecture and three PointNet++ variations. These variations differ in the grouping module used inside the set abstraction (SA) layer. Besides the single-scale grouping (SSG) and multi-scale grouping (MSG) modules proposed by Qi~et~al. in~\cite{PointNet++}, we developed and evaluated a so-called multi-form grouping (MFG) module which is introduced later in this section.

All of our PointNet models have been adapted for the usage with radar point clouds in terms of dimensions, i.e., two spatial and multiple feature dimensions. Moreover, we omitted sampling in the first SA layer of our PointNet++ models because the radar measurements are already sparse.

\subsection{Multi-Form Grouping} \label{section:mulit_form_group}

\begin{figure}[t]
    \centerline{
		\subfigure[Circular grouping.]{
		    \begin{tikzpicture}
		        \begin{axis}[
            		width=0.95*\linewidth, 			
            		grid=major, 					
            		grid style={dashed,gray!30}, 	
            		xlabel= $y$ in \si{\meter}, 	
            		ylabel= $x$ in \si{\meter},
            		x dir=reverse,
            		axis equal image,
            		xmin=-40,
            		xmax=40,
            		ymax=48,
            		ymin=-7,
            		colormap={CM}{
            			samples of colormap=(4 of radar_anomalies)},
            		colormap access=piecewise constant,
            		point meta min=0, point meta max=4,
            		legend columns = 3,
            		legend style={fill=white, fill opacity=0.8, draw opacity=1.0, text opacity=1.0, at={(0.925,0.175),anchor=south}}, 
            		/tikz/every even column/.append style={column sep=0.3cm},
            		legend cell align={left}
            		]             		\VisualizeRadar{resources/figures/angular_anomaly_fc_25_with_sensor_calib_applied.csv}{
            			0={mark=*, black},
            			4={mark=*, blue},
            			2={mark=*, red},
            			3={mark=*, black},
            			1={mark=*, orange}
            		}
            		\DrawBox{line width=0.35mm, gray}{-19}{-6.5}{25}{2.5}{6.5}
            		\addplot[color=brown!80!black, only marks, style={mark=*, fill=brown!20!black}, mark size=20, fill opacity=0.025] coordinates {(23.07, 22.01)};
            		\addlegendentry{stationary}
            		\addlegendentry{moving}
            		\addlegendentry{anomalous}
		        \end{axis}
		    \end{tikzpicture}
		    \vspace{-0.25cm}
		    \label{fig:multi_form_grouping:circle}
		}
	}
	\centerline{
		\subfigure[Ring grouping.]{
			\begin{tikzpicture}
		        \begin{axis}[
            		width=0.95*\linewidth, 			
            		grid=major, 					
            		grid style={dashed,gray!30}, 	
            		xlabel= $y$ in \si{\meter}, 	
            		ylabel= $x$ in \si{\meter},
            		x dir=reverse,
            		axis equal image,
            		xmin=-40,
            		xmax=40,
            		ymax=48,
            		ymin=-7,
            		colormap={CM}{
            			samples of colormap=(4 of radar_anomalies)},
            		colormap access=piecewise constant,	
            		point meta min=0, point meta max=4,
            		legend columns = 3,
            		legend style={fill=white, fill opacity=0.8, draw opacity=1.0, text opacity=1.0, at={(0.925,0.175),anchor=south}}, 
            		/tikz/every even column/.append style={column sep=0.3cm},
            		legend cell align={left}
            		]
            		\begin{scope}[on background layer]
            		\clip (-40, 0) rectangle (80, 50);
            		\addplot[color=brown!80!black, only marks, style={mark=*, fill=brown!20!black}, mark size=76, fill opacity=0.025, clip mode=individual, forget plot] coordinates {(0, 0)};
            		\addplot[color=brown!80!black, only marks, style={mark=*, fill=white}, mark size=60, clip mode=individual, forget plot] coordinates {(0, 0)};
            		\end{scope}            		\VisualizeRadar{resources/figures/angular_anomaly_fc_25_with_sensor_calib_applied.csv}{
            			0={mark=*, black},
            			4={mark=*, blue},
            			2={mark=*, red},
            			3={mark=*, black},
            			1={mark=*, orange}
            		}
            		\DrawBox{line width=0.35mm, gray}{-19}{-6.5}{25}{2.5}{6.5}
            		\addlegendentry{stationary}
            		\addlegendentry{moving}
            		\addlegendentry{anomalous}
		        \end{axis}
		    \end{tikzpicture}
		    \vspace{-0.25cm}
		    \label{fig:multi_form_grouping:ring}
	    }
	}
	\caption{Illustration of multi-form grouping of a radar measurement with the hypothesis that the anomalous radar targets (red) originate from the preceding vehicle (gray bounding box).\label{fig:multi_form_grouping}}
	\vspace{-0.25cm}
\end{figure}
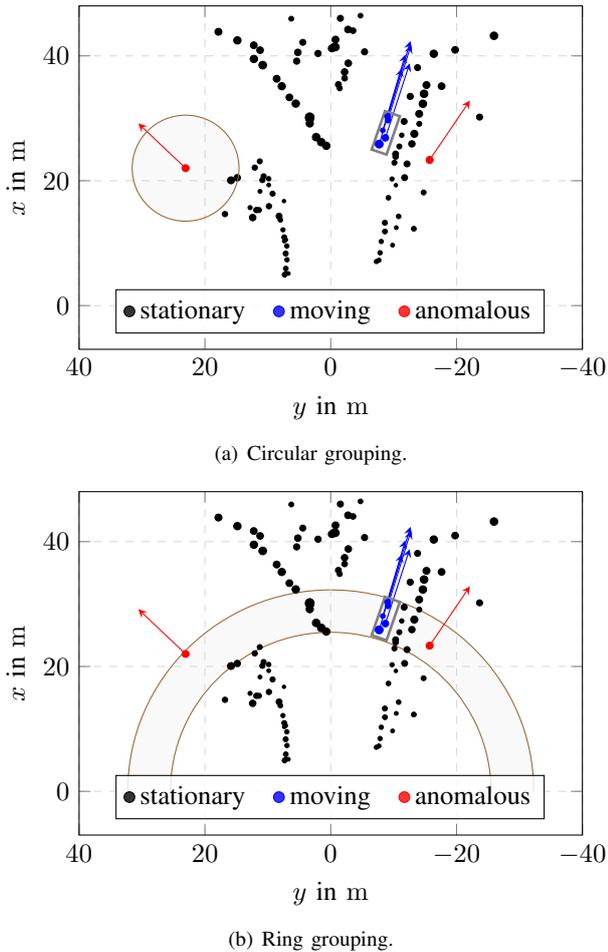
The multi-form grouping module is mainly motivated by~\cite{komarichev2019cnn} and \cite{sheshappanavar2020anovellocal}.
In~\cite{komarichev2019cnn} annularly CNNs are proposed on 3D point clouds where each point is utilized as center of annular convolutions.
Moreover, in~\cite{sheshappanavar2020anovellocal} the authors investigate the benefits of using ellipsoids instead of ball spheres for queried regions in PointNet++.
Rather than using an ellipsoid, we propose the usage of a ring with the origin as center.
To the best of our knowledge, this approach has not been investigated for PointNet++ architectures so far.

The reason for our choice is indicated in Fig.~\ref{fig:multi_form_grouping:ring}, as some of the considered anomalies occur in a ring-shaped region around the sensor origin within the same range as car targets.
The ring grouping is exemplarily illustrated for the targets of the preceding vehicle in Fig.~\ref{fig:multi_form_grouping:ring}.
In contrast to elliptical and circular regions, the ring-shaped querying can be easily realized by filtering the range of each target, especially if the spatial information of the radar data is represented using polar coordinates $(r, \phi)$. 
Besides that, a circular region that includes both the anomalous targets as well as the car targets covers a larger area than the corresponding ring, see Fig.~\ref{fig:multi_form_grouping:circle}. 
Because of that, the ring contains fewer targets than the circle and is thus more memory efficient. 
Nevertheless, we expect circular neighborhoods to be beneficial, e.g., for the kind of anomalies in Fig.~\ref{fig:considered_anomalies} which is assumed to be mainly related to the multi-path propagation effect.
Thus, we incorporate both querying forms into a single module, which leads to the naming of MFG. 
In addition to that, our module also includes the idea of using the neighborhood information of multiple different scales for both querying forms, as introduced with MSG. 
The components of the multi-form grouping module are depicted in Fig.~\ref{fig:multi_form_grouping}.

\section{Dataset}\label{section:dataset}

For training and testing of our anomaly detection methods, we created a hand-labeled dataset consisting of real-world radar data. The dataset has been recorded with the test vehicle of Ulm University~\cite{kunz2015autonomous}. The test vehicle is equipped with three front-facing ARS 408-21 radar sensors. One sensor is mounted on the center of the front bumper; the other two radars are mounted on the front corners of the vehicle. Since our objective is to use solely data of a single radar sensor for the detection of anomalies, this enables us to effectively create three datasets, one for each radar sensor.

The recorded sequence has a length of approximately \SI{3.5}{\minute} and represents a drive along an urban scenario in Ulm, Germany. More precisely, the chosen route includes a roundabout and several intersections. The sequence contains measurements of all three front radar sensors and images of the front camera. The measurements include both the ego-motion compensated and uncompensated Doppler velocity, as well as information about the ego-motion itself.

Furthermore, we limit the maximum range of the considered targets to \SI{70}{\meter}. This is necessary to ensure a correct ground truth labeling of radar anomalies based on the available camera images. Moreover, we only label anomalous targets with a significant Doppler velocity, although some of the effects discussed in Section~\ref{section:sensor_setup}, e.g., multi-path propagation, also apply to stationary targets. This is mainly motivated by the sparsity of the radar point clouds, which makes it almost impossible to distinguish between stationary anomalies and, for instance, a correct measurement of the ground.

In addition, it should also be noted that anomalies represent only approximately \SI{2}{\percent} of the total radar targets. Hence, our dataset is highly unbalanced, which affects the training process. We additionally investigated the distribution of anomalies across the measurements of the dataset. Thereby, we observed that approximately \SI{75}{\percent} of the radar point clouds contain at least one anomaly. Nevertheless, we also include measurements without any anomalies in our dataset.

\section{Experiments} \label{section:experiments}

In this section, we evaluate the proposed architectures on our real-world dataset. 

\subsection{Training} \label{section:training}

The training process has been performed solely using a main data part of the radar on the front center of the vehicle. 
This enables assessments of the generalization ability of our method when testing on data of the other two radars. 

To obtain batches for the training process, we randomly duplicate some of the radar targets during training in order to reach the same size of $250$ involved measurements in a batch. 
Furthermore, training of the models is performed with the Adam optimizer. We use a batch size of $48$ and learning rate scheduling. This schedule starts with a learning rate of $2 \times 10^{-4}$, which is halved every ten epochs. We train the models for $100$ epochs. Moreover, we apply data augmentation to avoid overfitting of our models. A challenging aspect for the training process is the unbalanced dataset. We address this problem by artificially increasing the number of anomalies in a radar measurement by combining them with the anomalies of consecutive measurements. To be more precise, for each measurement we consider the anomalies of the three measurements before and after. Each of these anomalies is inserted into the measurement with a probability of \SI{75}{\percent}. Both, the number of considered measurements and the probability are empirically chosen. Besides increasing the number of anomalies, we also scale the contribution of a target to the loss based on its class. More specifically, anomalies contribute nine times more to the loss than normal targets. This ratio is chosen empirically but was initially oriented at the ratio of normal targets and anomalies in the dataset. By increasing this ratio, we can increase the penalty of false negatives, i.e., non-detected anomalies. But it must be noted that this also decreases the penalty for false positives, which represent normal targets that have been classified as anomalies. Thus, a trade-off is necessary.

\subsection{Quantitative Results} \label{section:quant_results}

For the evaluation of our anomaly detection method, the $F_1$ score is used as metric. The $F_1$ score is the harmonic mean between precision $P$ and recall $R$ and is defined by
\begin{equation}
F_1 = \frac{2 P R}{P + R} .
\end{equation}

\sisetup{detect-weight=true,detect-inline-weight=math}
\begin{table}[t!]
	\caption{Results for the anomaly detection in radar data. Different models of the anomaly detection method are evaluated on our split dataset using $F_1$ score.}
	\vspace{-0.25cm}
	\label{tab:anomaly:other}
	\begin{center}
	\begin{tabular}{c|c|c|c|c}
		\toprule
		\diagbox[width=6.5em]{data}{model} & PointNet & \thead{PointNet++ \\ SSG} & \thead{PointNet++ \\ MSG} & \thead{PointNet++ \\ MFG} \\
		\midrule
		left    & \SI{66.07}{\percent} & \SI{66.89}{\percent} & \SI{74.04}{\percent}  & \bfseries{\SI{76.03}{\percent}}\\ 
		right    & \SI{77.01}{\percent} & \SI{73.12}{\percent} & \SI{81.97}{\percent} & \bfseries{\SI{82.21}{\percent}}\\
		center   & \SI{77.34}{\percent} & \SI{77.39}{\percent} & \SI{83.52}{\percent} & \bfseries{\SI{84.76}{\percent}} \\
		\midrule
		intersections  & \SI{51.89}{\percent} & \SI{54.55}{\percent} & \SI{59.94}{\percent} & \bfseries{\SI{63.62}{\percent}} \\
		\thead{without\\ intersections}  & \SI{74.74}{\percent} & \SI{72.59}{\percent} & \SI{80.87}{\percent} & \bfseries{\SI{81.64}{\percent}} \\
		\bottomrule
	\end{tabular}
	\end{center}
	\vspace{-0.6cm}
\end{table}
Table~\ref{tab:anomaly:other} shows the results of the four variations of the anomaly detection method in radar data.
First, we split the whole test dataset for evaluation purposes in data of the center, left, and right radar sensor. As already mentioned, we solely trained with data of the radar at the front center. For this reason, the test data of the center radar are rather small. For a more comprehensive evaluation and also the investigation of the generalization ability, data of the left and right sensor are used for testing. 
We notice that the PointNet model performs similar to the PointNet++ SSG model except for the right sensor data where the PointNet model is even better.
This indicates that the information gained from a single neighborhood by the SSG does not bring a notable benefit. 
Nevertheless, the PointNet++ MSG and MFG models, which consider the neighborhood of multiple scales, outperform the PointNet model.
More precisely, the PointNet++ MFG model performs best. This indicates that our modification of MSG with adapted grouping regions, which leads to the MFG module, is beneficial.
Beyond that, we observe for all models that the $F_1$ score on data of the front left radar is lower than on data of the front right radar.
This might be caused by the fact that the front left radar sensor is slightly oriented towards oncoming traffic, whereas the front right radar sensor is rotated into the other direction (right-hand traffic).
As a result of that, an oncoming vehicle, which may cause anomalies, is longer visible for the front left radar. 
Thereby, the number of anomalies is higher in the dataset of the front left and lower in the dataset of the front right radar sensor. 
As a result of that, the number of anomalies, which are challenging for the models, may also differ, leading to the differences observed in the results.

Second, we split the test data of the left and right radar in one part only with intersection scenarios and another part with everything else.
This is done because we hypothesize that the characteristics of the anomalies of our radar sensor in intersection scenarios, which are associated with slow ego-velocities, differ relevantly.
On closer inspection, we observed that the ego-velocity affects the characteristics of the anomalies significantly which is possibly caused by a transition of range gates and thus velocity ambiguities.
The general comparison results of the four different models are still similar.
However, the results support our hypothesis because the performance on intersection data is significantly worse than on data without intersections.
This is caused by the underrepresentation of these scenarios in our dataset, which makes them more challenging for our networks.

\begin{table}
	\caption{Mean inference time of the different models for  processing the anomaly detection method.}
	\vspace{-0.25cm}
	\label{tab:anomaly:inference}
	\begin{center}
	\begin{tabular}{c|c|c|c|c}
		\toprule
		\diagbox[width=6.9em]{measure}{model} & PointNet & \thead{PointNet++\\ SSG} & \thead{PointNet++\\ MSG} & \thead{PointNet++\\ MFG} \\
		\midrule
		
		\thead{inference\\ time}  & \bfseries \SI{1.6}{\milli\second} & \SI{13.4}{\milli\second} & \SI{26.7}{\milli\second}  & \SI{23.4}{\milli\second}\\ 
		\bottomrule
	\end{tabular}
	\end{center}
	\vspace{-0.5cm}
\end{table}
In Table~\ref{tab:anomaly:inference}, the inference times of the anomaly detection algorithms are illustrated. 
The tests were performed on a Linux workstation with a single \textit{NVIDIA GeForce RTX 2070 SUPER} GPU. 
It is worth mentioning that the computational performance was not the main focus of our work and is still perfectible.
We observe that the improved performance of the PointNet++ MSG and PointNet++ MFG models is achieved at the cost of a much higher time complexity in comparison to the PointNet++ SSG model.
However, it is worth mentioning that the inference for our PointNet++ MFG model is about \SI{3}{\milli\second} faster than for the original PointNet++ MSG model. 
This supports our claim that the ring-shaped query is more efficient than its circular pendant. 
Moreover, it should be noted that the PointNet++ SSG model has an inferior time complexity than the PointNet model, although both models achieved comparable $F_1$ scores.

\subsection{Qualitative Examples} \label{section:qual_examples}

\begin{figure*}[!t]
	\centerline{
	    \subfigure{
		    \hspace*{0.775cm}
		    \includegraphics[width=0.335\textwidth]{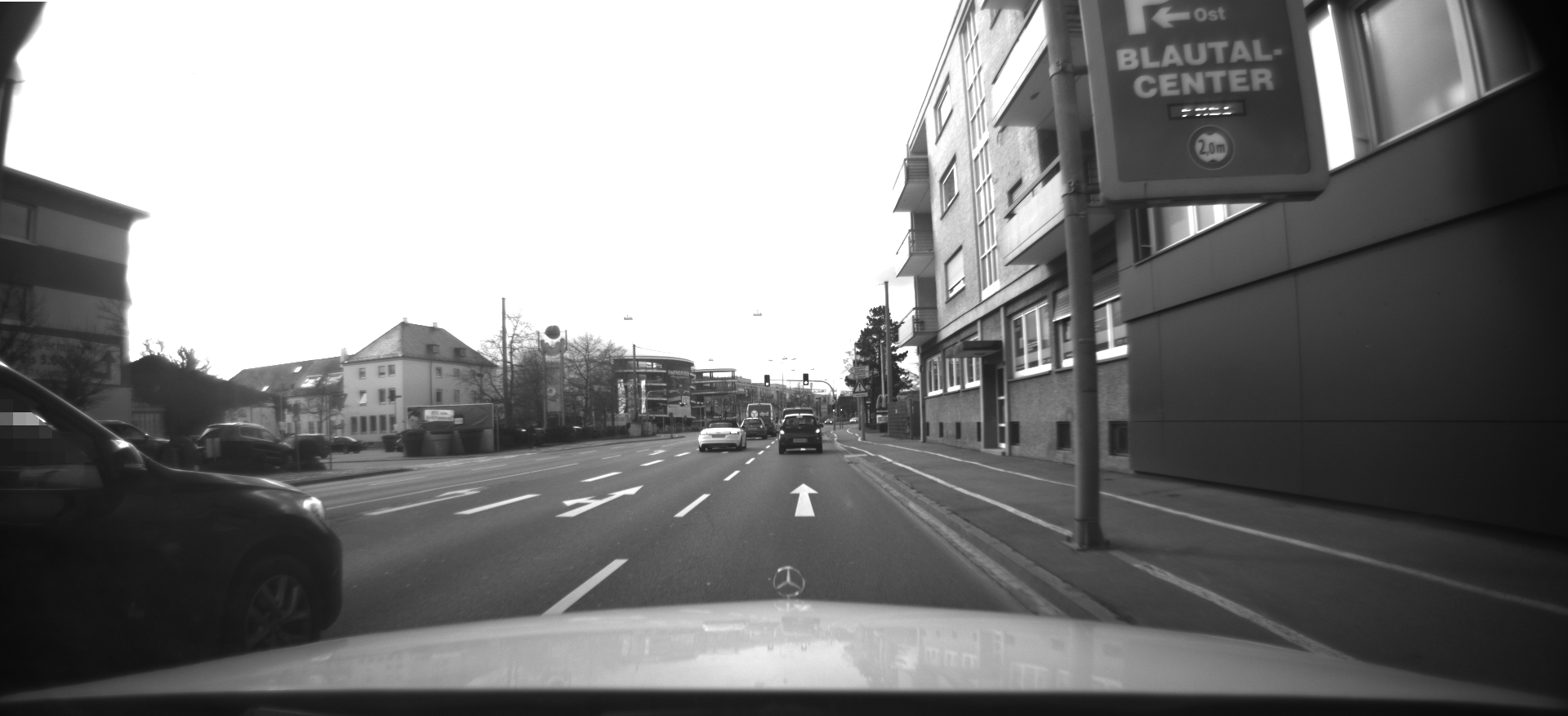}
	    }
	    \hfill
	    \subfigure{
		    \includegraphics[width=0.335\textwidth]{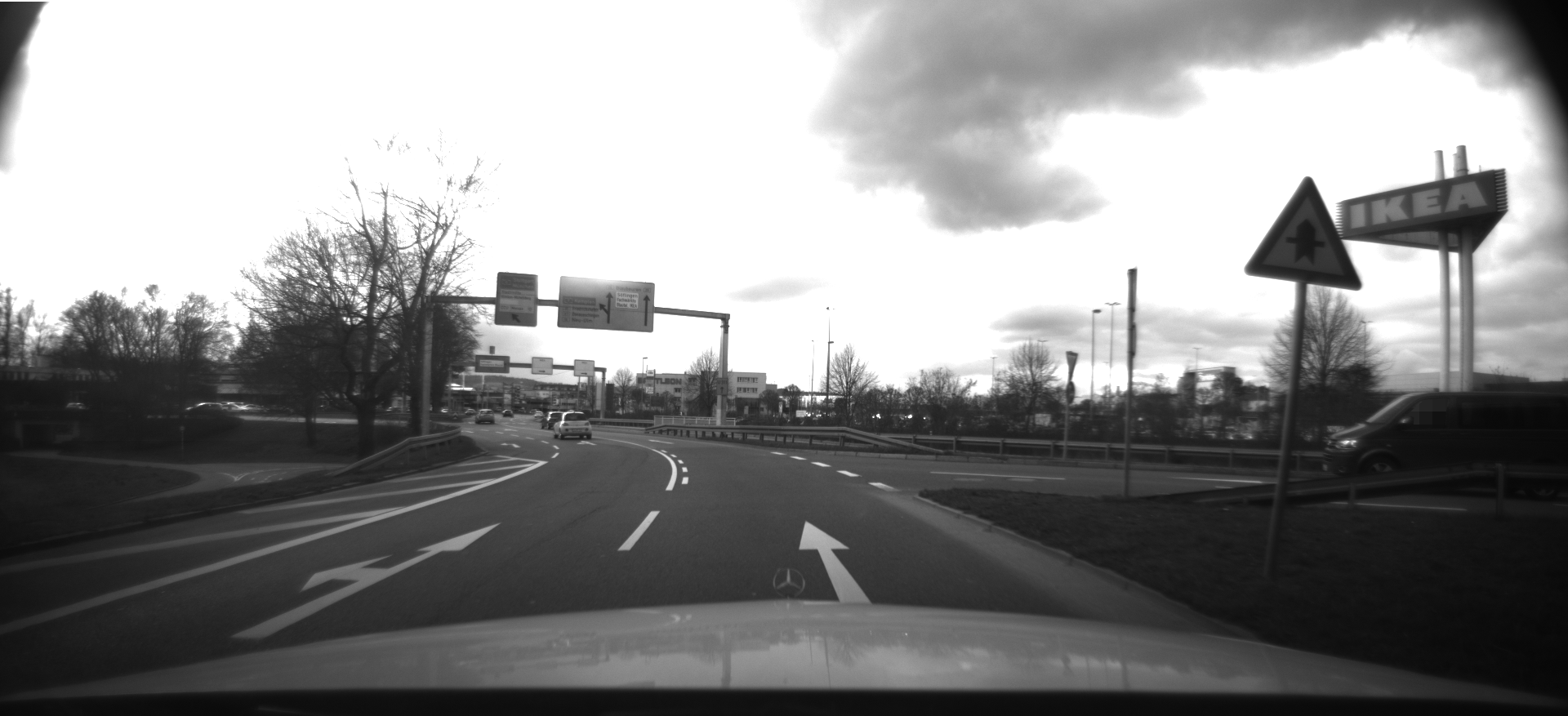}
	    }   
    }
	\addtocounter{subfigure}{-2}
    \centerline{
	    \subfigure[Exemplary illustration of a radar measurement near an intersection where our model classifies two anomalies with a low ego-motion compensated Doppler velocity incorrectly.]{
    		\begin{tikzpicture}
    			\begin{axis}[
    			grid=major, 					
    			grid style={dashed,gray!30}, 	
    			xlabel= $y$ in \si{\meter}, 	
    			ylabel= $x$ in \si{\meter},	
    			x dir=reverse,
    			ytick={0, 20, 40, 60, 80},
    			yticklabels={0,20,40,60,80},				
    			axis equal image,
    			xmin=-50,
    			xmax=50,
    			ymax=78,
    			ymin=-23,
    			colormap={CM}{
    				samples of colormap=(5 of radar_qualitative)},
    			colormap access=piecewise constant,
    			point meta min=0, point meta max=4,
    			legend style={fill=white, fill opacity=0.8, draw opacity=1.0, text opacity=1.0, at={(0.83,0.21),anchor=south}},
    			/tikz/every even column/.style={
    				column sep=0.3cm,
    			},
    			legend cell align={left},
    			legend columns = 2
    			]    			\VisualizeRadar{resources/figures/qualitative_mfg_fl_1800_with_sensor_calib_applied.csv}{
    				0={mark=*, black},
    				1={mark=*, green},
    				4={mark=*, blue},
    				3={mark=*, orange},
    				2={mark=*, red}
    			}
    			\DrawBox{line width=0.35mm, gray}{0}{4}{0}{3}{3};
    			\DrawBox{line width=0.35mm, gray}{0}{1.5}{23}{3}{6};
    			\DrawBox{line width=0.35mm, gray}{0}{5}{27.5}{3}{6};
    			\DrawBox{line width=0.35mm, gray}{0}{1.5}{36}{3}{7};
    			\DrawBox{line width=0.35mm, gray}{0}{4}{43}{3}{6};
    			\DrawBox{line width=0.35mm, gray}{0}{4}{52}{3}{8};
    			\DrawBox{line width=0.35mm, gray}{0}{3}{63}{2.5}{3};
    			\addlegendentry{TN (stationary)}
    			\addlegendentry{TP}
    			\addlegendentry{TN (moving)}
    			\addlegendentry{FN}
    			\end{axis}
    		\end{tikzpicture}
    		\label{fig:qualitative_examples:near_intersection}
	    }
	    \hfill
	    \subfigure[Exemplary illustration of a radar measurement with many correctly detected anomalies that are probably related to multi-path propagation involving the ego-vehicle.]{
        	\begin{tikzpicture}
        	    \begin{axis}[
        			grid=major, 					
        			grid style={dashed,gray!30}, 	
        			xlabel= $y$ in \si{\meter}, 	
        			ylabel= $x$ in \si{\meter},	
        			x dir=reverse,
        			ytick={0, 20, 40, 60, 80},
        			yticklabels={0,20,40,60,80},				
        			axis equal image,
        			xmin=-50,
        			xmax=50,
        			ymax=78,
        			ymin=-23,
        			colormap={CM}{
        				samples of colormap=(5 of radar_qualitative)},
        			colormap access=piecewise constant,
        			point meta min=0, point meta max=4,
        			legend style={fill=white, fill opacity=0.8, draw opacity=1.0, text opacity=1.0, at={(0.83,0.21),anchor=south}},
        			/tikz/every even column/.style={
        				column sep=0.3cm,
        			},
        			legend cell align={left},
        			legend columns = 2
        			]          			\VisualizeRadar{resources/figures/qualitative_mfg_fl_170_with_sensor_calib_applied.csv}{
        				0={mark=*, black},
        				1={mark=*, green},
        				4={mark=*, blue},
        				2={mark=*, red},
        				3={mark=*, orange}
        			}
        			\addlegendentry{TN (stationary)}
        			\addlegendentry{TP}
        			\addlegendentry{TN (moving)}
        			\DrawBox{line width=0.35mm, gray}{25}{13}{34}{3}{5};
        			\DrawBox{line width=0.35mm, gray}{25}{20}{52}{3}{5};
        			\DrawBox{line width=0.35mm, gray}{50}{-12}{9}{3}{4};
        			\end{axis}
        		\end{tikzpicture}
		        \label{fig:qualitative_examples:good}
		    }
    }
	\caption{Qualitative examples of predictions of our anomaly detection method. On the left, a radar measurement with erroneous prediction is presented, i.e., false negatives (FN). On the right, a radar measurement is depicted where all anomalies are correctly detected, i.e., true positives (TP).\label{fig:qualitative_examples}}
	\vspace{-0.25cm}
\end{figure*}
Qualitative examples of the anomaly detection prediction are visualized in Fig.~\ref{fig:qualitative_examples}.
An exemplary measurement which leads to wrong predictions is visualized in Fig.~\ref{fig:qualitative_examples:near_intersection}. 
The vehicles highlighted in this measurement are driving towards an intersection and therefore slowing down. 
Moreover, the measurement contains five anomalies.
Thereby, two of these anomalies have a significantly lower ego-motion compensated Doppler velocity than the remaining ones.
It is worth mentioning that the anomalies with low ego-motion compensated Doppler velocities are not detected, whereas our model correctly classifies the remaining ones as anomalies.
This also indicates that scenarios in which traffic slows down are challenging for our detector. 
Thus, the bad performance of our detector may be caused by the fact that scenarios with a significantly slower velocity of the ego-vehicle and other road users are underrepresented in our dataset.
Nevertheless, the quantitative evaluation already showed that our model is capable of detecting the majority of the anomalies. As a result of that, Fig.~\ref{fig:qualitative_examples:good} displays an exemplary measurement in which our model performs well. It is worth mentioning that the anomalies, which are probably related to multi-path propagation effects involving the ego-vehicle, tend to cluster in this measurement. As a consequence, the direct neighborhood of these anomalies contains more anomalous than normal targets. Although this makes the detection of anomalies more challenging, our model detects all of these and the other anomalies correctly.

\subsection{Discussion} \label{section:discussion}

In general, the results of the anomaly detector in radar data are promising. 
When taking inference time and $F_1$ score into account, the PointNet model is a good balance between these two aspects and also a suitable choice for systems with limited computational capacity. 
The PointNet++ MFG model provides significantly higher performance which is accompanied by higher computational requirements.
However, the inference time of our PointNet++ MFG model is lower than the original MSG model. 

Certainly, it is important to note that the used dataset is limited in several aspects. 
Firstly, the amount of data and variety of situations is limited. 
More precisely, only urban scenarios are covered in the dataset so far.
Besides, the three subdatasets, one for every sensor, are correlated because, although the data are from different sensors, they still cover the same driving sequence.
In addition, the variety of objects is limited, e.g., trucks are highly underrepresented. 
However, these limitations can be overcome by extending the dataset.

Moreover, we restricted ourselves to use only a single radar measurement for the anomaly detection.
In this way, the anomalies can be neither detected using the temporal information of multiple consecutive measurements nor by the fusion of overlapping point clouds obtained from different radars. 
However, this enables us to use a classical centralized fusion approach where temporal information is not taken into account until the tracking stage.
In addition, this approach facilitates us to combine our proposed anomaly detection with a single-shot object detector operating also on single radar measurements such as~\cite{griebel2019Car}.

\section{Conclusion} \label{section:conclusion}

In this work, we tackle the problem of anomaly detection in radar data using a single radar measurement. 
We first described and defined the anomalies which we want to detect in our real-word data and hypothesize the reasons for these anomalies. 
In doing so, we observed that approximately \SI{75}{\percent} of the radar point clouds in our dataset contain at least one anomaly. 
For the anomaly detection, we used the PointNet architecture family as a base. 
Thereby, we proposed a novel grouping algorithm for the PointNet++ architecture, the multi-form grouping. 
In contrast to classical circular grouping, our approach takes the characteristics of anomalous radar targets into account. 
This enables us to outperform the reference implementation with circular grouping both in terms of the $F_1$ score as well as the inference time. 
Overall, our approach shows promising results for detecting anomalies in radar data.

In our future work, we aim to extend this approach also for other kind of anomalies occurring in radar measurements. 
To this end, the dataset should be increased in size and various additional real-world scenarios should be included, e.g., non-urban scenarios with a speed limit higher than \SI{50}{\kilo\meter\per\hour}.
Besides that, the combination and interaction of our proposed anomaly detection with a single-shot object detector operating on radar measurements such as~\cite{griebel2019Car} should be further investigated. 
Additionally, generative adversarial networks (GANs) seem suitable for further investigations of anomaly detection in radar data.
Finally, a comparison to other radar sensors regarding anomalies and its detection should be explored.



%


\bibliographystyle{IEEEtran}
\bibliography{references}

\end{document}